%% file: paper.tex
\documentclass[10pt]{article}

\usepackage[margin=0.85in]{geometry}
\usepackage[T1]{fontenc}
\usepackage{enumitem}
\usepackage{float}
\usepackage{flushend}
\usepackage{graphicx}
\usepackage{listings}
\usepackage{microtype}
\usepackage{needspace}
\usepackage[numbers,sort&compress]{natbib}
\usepackage{placeins}
\usepackage[table]{xcolor}
\usepackage{tabularx}
\usepackage{tikz}
\usepackage[hidelinks]{hyperref}

\usetikzlibrary{arrows.meta}
\graphicspath{{figures/}}
\setlist{nosep,leftmargin=*}
\definecolor{AlbBlue}{HTML}{2563EB}
\definecolor{AlbGreen}{HTML}{16A34A}
\definecolor{AlbOrange}{HTML}{D97706}
\definecolor{AlbPurple}{HTML}{7C3AED}
\definecolor{AlbRed}{HTML}{DC2626}
\definecolor{AlbGray}{HTML}{4B5563}

\newcommand{\code}[1]{\nolinkurl{#1}}

\newcommand{\tablebodycolors}{\rowcolors{2}{white}{gray!5}}
\newcolumntype{Y}{>{\raggedright\arraybackslash}X}

\lstdefinestyle{python}{
  language=Python,
  basicstyle=\ttfamily\footnotesize,
  keywordstyle=\color{AlbBlue}\bfseries,
  commentstyle=\color{AlbGray},
  stringstyle=\color{AlbGreen!70!black},
  showstringspaces=false,
  columns=fullflexible,
  keepspaces=true,
  breaklines=true,
  frame=single,
  rulecolor=\color{gray!35},
  backgroundcolor=\color{gray!3},
  xleftmargin=2pt,
  xrightmargin=2pt,
  aboveskip=5pt,
  belowskip=5pt
}

\title{AlbumentationsX: One Augmentation Pipeline for Images and Related Annotations}

\author{
Vladimir Iglovikov\\
Albumentations LLC\\
\texttt{vladimir@albumentations.ai}\\
\url{https://github.com/albumentations-team/AlbumentationsX}
}

\date{}

\begin{document}
\maketitle

\begin{abstract}
Augmentation can corrupt a training example when an image and its annotations receive different random changes. A crop must use the same coordinates for the image, mask, boxes, keypoints, stereo views, video frames, or volume. Code paths that choose these values separately can silently misalign the data.

AlbumentationsX keeps the transform list, probabilities, annotation settings, and random seed in one \code{Compose} object. Each call chooses random values once and applies them to every supported part of the training example. The library keeps each object's mask, box, and label together and lets projects add their own transforms. It can also save the pipeline definition, show what happened in one call, and run that call again.

The examples place \code{Compose} after files have been decoded into arrays and before PyTorch groups examples into a batch. AlbumentationsX executes the declared transforms. Practitioners still decide whether a flip, crop, color change, or other operation preserves the correct label for their task.
\end{abstract}

\section{Introduction}

A crop applied to an image but not its mask corrupts a training example. The same failure appears when a box, keypoint, stereo view, video frame, or volume receives different crop, rotation, or resize values. Every related value must receive the same geometric change.

Data augmentation reuses labeled examples by changing their appearance or geometry during training. A brightness transform can simulate different lighting. A crop changes which part of the scene is visible and where objects appear. Blur can simulate motion or a camera that missed focus. Collecting and labeling representative real data remains the stronger answer when those examples can be obtained.

Every augmentation makes an assumption about the task: the change should leave the correct answer unchanged. A horizontal flip may be valid for road-scene segmentation because the same objects remain in the image. The same flip is usually wrong for optical character recognition because it mirrors the text. The dataset owner must choose transformations that match plausible variation and preserve the label.

A computer-vision training example often contains more than an image and one class label. This paper calls the complete example a \emph{sample}. Semantic segmentation uses an image and a pixel mask. Object detection uses an image, bounding boxes, and one class label per box. Pose estimation adds keypoints, such as the left wrist and right knee. A video sample contains an ordered sequence of frames. A 3D sample may contain a volume and a 3D mask.

AlbumentationsX calls each named value that a transform knows how to handle a \emph{target}. Images, masks, bounding-box arrays, and keypoint arrays are targets. In semantic segmentation, the image and mask are two targets in one sample. After a crop, the mask must still cover the same pixels as the object in the image.

An augmentation \emph{policy} is the transform list plus the rules that control it: order, probabilities, target settings, and random seed. AlbumentationsX stores this policy in one \code{Compose} object~\cite{albumentationsx,albumentationsdocs}. One call chooses the random values and applies them to every target in the sample. A crop therefore uses one set of coordinates for the image, mask, boxes, keypoints, related images, and every frame in a video clip.

A training input path usually reads a file, decodes it into arrays, augments one sample, and then groups several samples into a batch. AlbumentationsX performs the augmentation step while the image and annotations are still separate named values. The following sections show how to build one policy, keep related targets aligned, add a project-specific operation, and choose the record needed to inspect or repeat a random call.

\section{Build One Pipeline}

The example below crops every sample, flips half of them, and adjusts brightness and contrast or applies a gamma curve in 40\% of calls. It then normalizes the image for the model. The argument \code{p} is the probability that a transform runs; \code{p=1.0} means always. \code{OneOf} first decides whether its group runs and then chooses one child. In \code{BboxParams}, \code{coord_format="pascal_voc"} means that each box is stored as \code{[x_min, y_min, x_max, y_max]}. The setting \code{label_fields=["labels"]} keeps each surviving box paired with its class label after a crop. The value \code{seed=137} starts the policy's own random sequence.

\noindent\hspace*{4pt}\begin{minipage}{\dimexpr\columnwidth-8pt\relax}
\begin{lstlisting}[style=python]
import albumentations as A

policy = A.Compose(
    [
        A.RandomCrop(height=384, width=512, p=1.0),
        A.HorizontalFlip(p=0.5),
        A.OneOf(
            [
                A.RandomBrightnessContrast(p=0.5),
                A.RandomGamma(p=0.5),
            ],
            p=0.4,
        ),
        A.Normalize(p=1.0),
    ],
    bbox_params=A.BboxParams(
        coord_format="pascal_voc",
        label_fields=["labels"],
    ),
    seed=137,
)

sample = policy(
    image=image,
    mask=mask,
    bboxes=boxes,
    labels=labels,
)
\end{lstlisting}
\end{minipage}

The two children inside \code{OneOf} have the same \code{p}, so the group chooses each one equally often. \code{Normalize(p=1.0)} runs last, after the random image changes.

A fixed transform list is enough for a simple policy. Larger policies must also express which steps are optional, which alternative to choose, and whether order is fixed or random. Table~\ref{tab:policy-language} maps each decision to the API object that expresses it.

\begin{table}[H]
\centering
\footnotesize
\setlength{\tabcolsep}{3pt}
\renewcommand{\arraystretch}{1.16}
\tablebodycolors
\begin{tabularx}{\columnwidth}{@{}p{0.31\columnwidth}p{0.25\columnwidth}Y@{}}
\textbf{Need} & \textbf{API} & \textbf{Result} \\
\hline
Ordered steps & \code{Compose} & Run the declared sequence \\
Optional step & Transform \code{p} & Run it with probability \code{p} \\
Choose one & \code{OneOf} & Choose one child at random \\
Choose several & \code{SomeOf} & Choose a requested number at random \\
Choose and reorder & \code{RandomOrder} & Choose a subset, then apply it in random order \\
Grouped steps & \code{Sequential} & Nest one ordered block \\
\end{tabularx}
\caption{These API objects keep order, optional steps, and random choices inside the policy.}
\label{tab:policy-language}
\end{table}

The \code{Compose} seed controls this policy's sequence of random choices; global \code{random} and \code{numpy.random} seeds do not control it. Concurrent calls use separate random generators, so one call cannot consume another call's random numbers. The declared order remains visible: crop first, change appearance second, and normalize last. Section~5 shows how to repeat one sample without depending on call order.

\Needspace{8\baselineskip}
\section{Keep Each Object with Its Mask and Box}

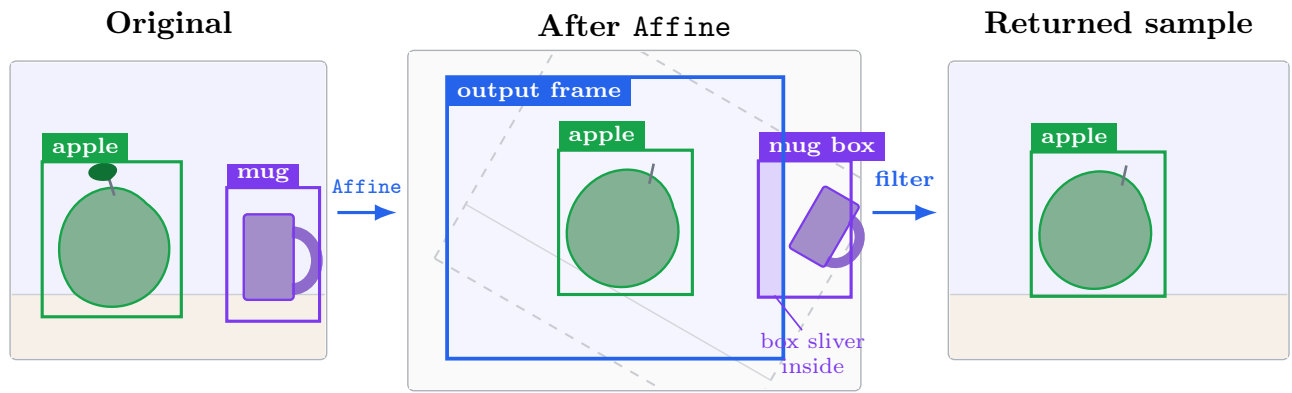
\begin{figure}[htbp]
\centering
\begin{minipage}{0.98\textwidth}
  \input{figures/instance-affine-diagram.tex}
\end{minipage}
\caption{After the affine transform, the mug's mask is outside the blue output frame, but its axis-aligned box still overlaps the frame by a narrow strip. AlbumentationsX 2.4.0 uses the empty mask to remove the mug's mask, box, and class label together. The apple remains with aligned annotations.}
\label{fig:instance-affine}
\end{figure}

Instance segmentation predicts each object separately. One object can have a mask, a bounding box, a class label, and keypoints. A geometric transform must update all of these annotations together. An object near the image edge can move completely outside the returned image after rotation and translation. Its returned box still has horizontal and vertical sides; this is an \emph{axis-aligned} box. That box may overlap the image by a narrow strip even when the transformed mask has no pixels inside it. Figure~\ref{fig:instance-affine} shows why the empty mask must remove the complete object.

AlbumentationsX provides this behavior through \code{instances}: a list with one dictionary per object. Each dictionary keeps that object's mask, box, and class label together. The \code{bbox_labels} dictionary holds the box fields named by \code{BboxParams.label_fields}. The value \code{instance_binding=["masks", "bboxes"]} selects masks and boxes for joint processing. \code{Compose} collects each object's \code{mask} and \code{bbox}, transforms them, and returns them to the same object dictionary.

The example fixes the rotation at 30 degrees and the horizontal movement at 52 pixels. The values \code{fill=0} and \code{fill_mask=0} fill newly exposed image and mask pixels with zero. The box filters keep any box with at least one pixel of area, including the mug's narrow strip.

\begin{lstlisting}[style=python]
instances = [
    {
        "mask": apple_mask,
        "bbox": apple_box,
        "bbox_labels": {"class_name": "apple"},
    },
    {
        "mask": mug_mask,
        "bbox": mug_box,
        "bbox_labels": {"class_name": "mug"},
    },
]
\end{lstlisting}

The policy binds those fields and applies the fixed affine transform:

\noindent\hspace*{4pt}\begin{minipage}{\dimexpr\columnwidth-8pt\relax}
\begin{lstlisting}[style=python]
instance_policy = A.Compose(
    [
        A.Affine(
            rotate=(-30, -30),
            translate_px={"x": (52, 52), "y": (0, 0)},
            fill=0,
            fill_mask=0,
            p=1.0,
        ),
    ],
    bbox_params=A.BboxParams(
        coord_format="pascal_voc",
        label_fields=["class_name"],
        min_area=1,
        min_visibility=0.0,
    ),
    instance_binding=["masks", "bboxes"],
    seed=137,
)

sample = instance_policy(
    image=image,
    instances=instances,
)
\end{lstlisting}
\end{minipage}

After the transform, the mug's mask is empty, so AlbumentationsX removes the mug dictionary from \code{sample["instances"]}. The apple stays in the same per-object format with its transformed mask, box, and label.

Other tasks use the same idea without the per-object container. Semantic segmentation passes \code{image=} and \code{mask=}. Detection and pose add \code{bboxes=} and \code{keypoints=}. A video clip uses \code{images=frames}; 3D data uses \code{volume=} and, when present, \code{mask3d=}.

\subsection{Keep stereo images and depth maps aligned}

Stereo vision uses two images of the same scene, one from the left camera and one from the right. Every allowed crop, resize, or other geometric transform must use the same values for both views. An RGB-D sample also contains a depth map, which stores the distance at each pixel. The depth map needs the same geometry as the color image, but it must skip brightness and contrast changes.

The setting \code{additional_targets={"right_image": "image"}} tells \code{Compose} to process \code{right_image} like the main image: both receive the same geometry and the same image-only transforms. The depth map is passed as \code{mask}, so it receives the shared geometry but skips image-only transforms. The code renames that output back to \code{depth} after the call.

Depth contains continuous measurements, so \code{mask_interpolation=cv2.INTER_LINEAR} selects linear interpolation. A segmentation mask contains class IDs and normally uses nearest-neighbor interpolation so resize does not invent new class values.

\noindent\hspace*{4pt}\begin{minipage}{\dimexpr\columnwidth-8pt\relax}
\begin{lstlisting}[style=python]
import cv2

paired_policy = A.Compose(
    [
        A.Resize(height=384, width=512, p=1.0),
        A.RandomCrop(height=352, width=480, p=1.0),
        A.RandomBrightnessContrast(p=0.3),
        A.Normalize(p=1.0),
    ],
    additional_targets={
        "right_image": "image",
    },
    mask_interpolation=cv2.INTER_LINEAR,
    seed=137,
)

paired = paired_policy(
    image=left_image,
    right_image=right_image,
    mask=depth_map,
)
paired["depth"] = paired.pop("mask")
\end{lstlisting}
\end{minipage}

Both color images receive the same resize and crop, and both are normalized. The depth map receives the resize and crop but skips brightness, contrast, and normalization. The \code{mask_interpolation} setting applies to every value processed as a mask in this \code{Compose}. If one sample contains both continuous depth and a class-ID mask, one interpolation setting is not enough. Use a custom transform that resizes depth with linear interpolation and the segmentation mask with nearest-neighbor interpolation.

Some stereo training methods need the same geometry but independent sensor noise in the two cameras. First use one pipeline for the shared geometry. Then use a separate pipeline for each view to change pixel values without moving pixels.

\subsection{Apply one crop to the whole video clip}

A video clip is an ordered sequence of frames. Cropping every frame independently would make the crop window jump around, creating artificial camera motion. Pass the complete video clip as \code{images=frames}. AlbumentationsX then uses the same crop and flip for every frame.

\noindent\hspace*{4pt}\begin{minipage}{\dimexpr\columnwidth-8pt\relax}
\begin{lstlisting}[style=python]
clip_policy = A.Compose(
    [
        A.RandomCrop(height=224, width=224, p=1.0),
        A.HorizontalFlip(p=0.5),
        A.RandomBrightnessContrast(p=0.3),
        A.Normalize(p=1.0),
    ],
    seed=137,
)

augmented_clip = clip_policy(
    images=frames,
)["images"]
\end{lstlisting}
\end{minipage}

Use \code{image=} for one multichannel image, such as a 9-channel satellite image. The channel count does not guarantee that every transform supports the array; some color transforms accept only one or three channels. Use \code{volume=} for a 3D volume and \code{mask3d=} for its 3D mask.

\FloatBarrier

\section{Add a Project-Specific Transform}

A transform catalog cannot cover every camera and dataset. A project may need to simulate a specific gain error, stripe pattern, or sensor defect. A custom transform puts that operation inside the same \code{Compose} pipeline as the built-in transforms, so it uses the same probability and seed rules.

The example below simulates sensor gain by multiplying every pixel by one number between 0.8 and 1.2. No pixel moves, so masks, boxes, and keypoints do not change. The class therefore derives from \code{ImageOnlyTransform}.

The \code{sample_parameters} method chooses the gain once for the call. The \code{apply} method receives that chosen number and performs only the pixel calculation. \code{Compose} supplies \code{SamplingContext}, whose random generators follow the policy seed.

\noindent\hspace*{4pt}\begin{minipage}{\dimexpr\columnwidth-8pt\relax}
\begin{lstlisting}[style=python]
import numpy as np
import albumentations as A
from albumentations.core.invocation import SamplingContext

class RandomGain(A.ImageOnlyTransform):
    def __init__(self, gain=(0.8, 1.2), p=0.5):
        super().__init__(p=p)
        self.gain = gain

    def sample_parameters(
        self, params, data, sampling: SamplingContext
    ):
        value = sampling.py_random.uniform(*self.gain)
        return {"gain": value}

    def apply(self, img, gain, **params):
        scaled = img.astype(np.float32) * gain
        return np.clip(scaled, 0, 255).astype(img.dtype)
\end{lstlisting}
\end{minipage}

Use \code{sampling.py_random} when the transform needs a few random numbers. Use \code{sampling.random_generator} when it needs a random array. Keep every random choice in \code{sample_parameters}; \code{apply} should only use the values it receives.

Choose the base class by the values that the operation must change. Table~\ref{tab:extension-paths} maps each project need to the corresponding API.

\begin{table}[H]
\centering
\footnotesize
\setlength{\tabcolsep}{3pt}
\renewcommand{\arraystretch}{1.16}
\tablebodycolors
\begin{tabularx}{\columnwidth}{@{}p{0.24\columnwidth}p{0.31\columnwidth}Y@{}}
\textbf{Project need} & \textbf{Start from} & \textbf{What changes} \\
\hline
Pixel or sensor effect & \code{ImageOnlyTransform} & Image values; spatial targets stay fixed \\
Spatial warp & \code{DualTransform} & Image, masks, boxes, and keypoints \\
3D volume operation & \code{Transform3D} & Volume and matching 3D targets \\
Project-specific input name & \code{CustomTransformsApplyMixin} & \code{apply\_to\_<key>} receives the same chosen parameters \\
\end{tabularx}
\caption{Choose the custom-transform base class by the kind of data the new operation changes.}
\label{tab:extension-paths}
\end{table}

First test the numerical operation without randomness; for example, verify that a pixel value of 100 becomes 120 when the gain is 1.2.

Then place the transform in a fixed-seed \code{Compose} call and check every affected input. The project supplies the sensor- or task-specific calculation and reuses the surrounding pipeline code.

\section{Repeat or Inspect a Random Call}

When an augmented sample looks wrong, first decide what must be repeated. A policy seed repeats the whole sequence: a new \code{Compose} object with the same policy and seed makes the same random choices when samples arrive in the same order. If the order changes, a sample may receive a different crop or color change.

Use \code{invocation_seed} when one sample must receive the same choices regardless of which other calls run first. Pass a stable integer with that sample:

\noindent\hspace*{4pt}\begin{minipage}{\dimexpr\columnwidth-8pt\relax}
\begin{lstlisting}[style=python]
sample = policy(
    image=image,
    mask=mask,
    invocation_seed=1_000_001,
)
\end{lstlisting}
\end{minipage}

A seed does not show which crop coordinates an earlier call used. Save the \code{Compose} definition when the goal is to rebuild the policy later. Set \code{save_applied_params=True} to record which transforms ran and the settings they exposed. Use \code{ReplayCompose} when the exact crop, flip decision, and other random values from one call must run again.

\begin{table}[H]
\centering
\footnotesize
\setlength{\tabcolsep}{3pt}
\renewcommand{\arraystretch}{1.18}
\tablebodycolors
\begin{tabularx}{\columnwidth}{@{}p{0.29\columnwidth}Y@{}}
\textbf{Question} & \textbf{Record to keep} \\
\hline
Repeat the ordered sequence? & Policy definition, seed, and sample order \\
Repeat one sample despite call order? & Policy definition and \code{invocation_seed} \\
Rebuild the policy later? & Saved \code{Compose} definition and library version \\
See which transforms ran? & \code{save_applied_params=True} output and sample identifier \\
Run the exact random call again? & \code{ReplayCompose} record for supported transforms \\
Repeat the complete loader run? & Sample order, framework seed, worker count, and whether workers persist \\
\end{tabularx}
\caption{Keep the smallest record that answers the question.}
\label{tab:saved-state}
\end{table}

Table~\ref{tab:saved-state} gives the minimum record for each question. A complete data-loading run also depends on the training framework's random state, the number of \code{DataLoader} workers, and whether those workers remain alive between epochs. The compact \code{save_applied_params} output is intended for inspection and may omit runtime values that \code{ReplayCompose} needs.

\section{Call AlbumentationsX from the PyTorch Dataset}

In PyTorch, \mbox{\texttt{Dataset.\_\_getitem\_\_}} usually prepares one sample, and \code{DataLoader} later groups returned samples into batches. Call AlbumentationsX inside \code{Dataset.\_\_getitem\_\_}, after JPEG or other file bytes have become arrays. At that point, the image and annotations still have separate names, so \code{Compose} can update them together. The example below returns a fixed-size image and mask.

\noindent\hspace*{4pt}\begin{minipage}{\dimexpr\columnwidth-8pt\relax}
\begin{lstlisting}[style=python]
import albumentations as A
from albumentations.pytorch import ToTensorV2
from torch.utils.data import Dataset

train_policy = A.Compose(
    [
        A.RandomCrop(height=384, width=512, p=1.0),
        A.HorizontalFlip(p=0.5),
        A.RandomBrightnessContrast(p=0.3),
        A.Normalize(p=1.0),
        ToTensorV2(p=1.0),
    ],
    seed=137,
)

class SegmentationDataset(Dataset):
    def __getitem__(self, index):
        image, mask = self.load_decoded(index)
        sample = train_policy(
            image=image,
            mask=mask,
        )
        return sample["image"], sample["mask"].long()
\end{lstlisting}
\end{minipage}

This placement keeps the image and mask together until the last geometric operation. Image-only transforms receive only the image. Normalization and \code{ToTensorV2} then produce the PyTorch tensors expected by the model. Because the crop returns a fixed size, PyTorch's default batching step can stack the images and masks. Detection pipelines may need a custom batching function because different images can contain different numbers of boxes.

Declare training and evaluation as separate policies. A validation policy usually contains deterministic resizing, padding, normalization, and tensor conversion. The two objects then answer two distinct review questions: which random variation trains the model, and which deterministic preprocessing defines evaluation?

\section{Related Libraries}

TorchVision v2, Kornia, and NVIDIA DALI can run supported image augmentations on GPUs; DALI can include decoding in the same pipeline~\cite{torchvisiontransforms,riba2020kornia,korniaaugmentation,dali}. In the cited releases, the public APIs expose 121 concrete 2D transform classes in AlbumentationsX, 61 in Kornia, and 38 in TorchVision v2~\cite{albumentationsx,korniaaugmentation,torchvisiontransforms}. The count excludes base classes, pipeline and policy containers, format-conversion helpers, and transforms limited to 3D data or spectrograms. AlbumentationsX also keeps geometric changes aligned across images and related annotations.

\section{Limits and Availability}

AlbumentationsX cannot infer whether a transform preserves the task label. Rotation may be valid for aerial imagery and harmful for a digit classifier. Color changes may simulate different lighting, but they can remove the signal used by a model that classifies fruit ripeness. Practitioners make these decisions from knowledge of the task and deployment data.

Target support varies by transform. For example, a color transform may accept an RGB image but reject a 9-channel array. Some arrays describe the camera rather than image pixels. A depth map can share the image's crop and resize. A camera calibration matrix stores camera geometry, so a crop or resize must update its numbers with a separate rule.

The examples use AlbumentationsX release 2.4.0 (commit \code{e85171bc777f}). The public repository is available under AGPL-3.0-only. The AGPL permits commercial use subject to its terms. Albumentations, LLC also offers separately negotiated commercial licenses with alternative rights defined by the applicable agreement. The source and documentation are available at the cited URLs~\cite{albumentationsx,albumentationsdocs}.

\section{Conclusion}

AlbumentationsX stores the transform list and its rules in one \code{Compose} object. Each call chooses random values once and applies them to the image and its related annotations. The \code{instances} input keeps each object's mask, box, label, and keypoints together. The same pipeline can process stereo pairs, depth maps, video clips, 3D volumes, and project-specific transforms. It can also save the policy, show which transforms ran, or repeat one exact call.

Use AlbumentationsX when a project needs one reviewable Python policy for decoded images and their related targets.

\scriptsize
\bibliographystyle{plainnat}
\bibliography{references}

\end{document}

%% file: figures/instance-affine-diagram.tex
\resizebox{\linewidth}{!}{%
\begin{tikzpicture}[
  font=\small,
  panel/.style={draw=AlbGray!45, rounded corners=2pt, fill=gray!3,
    line width=0.45pt},
  outputframe/.style={draw=AlbBlue, line width=1.2pt},
  flow/.style={-{Latex[length=2.5mm]}, draw=AlbBlue, line width=1.0pt},
  applebox/.style={draw=AlbGreen, line width=1.0pt},
  mugbox/.style={draw=AlbPurple, line width=1.0pt},
  applelabel/.style={fill=AlbGreen, text=white, font=\scriptsize\bfseries,
    inner xsep=3pt, inner ysep=1.4pt, anchor=south west},
  muglabel/.style={fill=AlbPurple, text=white, font=\scriptsize\bfseries,
    inner xsep=3pt, inner ysep=1.4pt, anchor=south west}
]
  \node[font=\normalsize\bfseries] at (1.78,4.30) {Original};
  \node[font=\normalsize\bfseries] at (7.00,4.30) {After \code{Affine}};
  \node[font=\normalsize\bfseries] at (12.43,4.30) {Returned sample};

  \draw[panel] (0,0.55) rectangle (3.55,3.90);
  \fill[blue!5] (0.02,1.28) rectangle (3.53,3.88);
  \fill[brown!12] (0.02,0.57) rectangle (3.53,1.28);
  \draw[AlbGray!28, line width=0.45pt] (0.02,1.28) -- (3.53,1.28);

  \path[fill=gray!55, draw=gray!70]
    (0.72,1.30)
    .. controls (0.45,1.62) and (0.50,2.18) .. (0.88,2.40)
    .. controls (1.12,2.54) and (1.36,2.48) .. (1.53,2.30)
    .. controls (1.90,2.03) and (1.84,1.50) .. (1.53,1.27)
    .. controls (1.29,1.09) and (0.94,1.10) .. cycle;
  \path[fill=AlbGreen, fill opacity=0.34, draw=AlbGreen, line width=0.75pt]
    (0.72,1.30)
    .. controls (0.45,1.62) and (0.50,2.18) .. (0.88,2.40)
    .. controls (1.12,2.54) and (1.36,2.48) .. (1.53,2.30)
    .. controls (1.90,2.03) and (1.84,1.50) .. (1.53,1.27)
    .. controls (1.29,1.09) and (0.94,1.10) .. cycle;
  \draw[AlbGray!80, line width=0.9pt] (1.17,2.39) -- (1.09,2.64);
  \path[fill=AlbGreen!70!black]
    (1.11,2.56) .. controls (0.84,2.52) and (0.79,2.72) .. (1.04,2.76)
    .. controls (1.20,2.76) and (1.27,2.67) .. cycle;
  \draw[applebox] (0.36,1.03) rectangle (1.92,2.77);
  \node[applelabel] at (0.36,2.77) {apple};

  \path[fill=gray!55, draw=gray!70, rounded corners=1.2pt]
    (2.62,1.22) rectangle (3.18,2.18);
  \draw[gray!70, line width=3.7pt]
    (3.16,1.98) arc[start angle=90,end angle=-90,x radius=0.28cm,y radius=0.32cm];
  \path[fill=AlbPurple, fill opacity=0.34, draw=AlbPurple, line width=0.75pt,
    rounded corners=1.2pt] (2.62,1.22) rectangle (3.18,2.18);
  \draw[AlbPurple, opacity=0.55, line width=3.7pt]
    (3.16,1.98) arc[start angle=90,end angle=-90,x radius=0.28cm,y radius=0.32cm];
  \draw[mugbox] (2.43,0.98) rectangle (3.47,2.48);
  \node[muglabel] at (2.43,2.48) {mug};

  \draw[flow] (3.67,2.20) -- (4.34,2.20);
  \node[font=\scriptsize\bfseries, text=AlbBlue, align=center]
    at (4.00,2.50) {\code{Affine}};

  \draw[panel] (4.46,0.20) rectangle (9.54,4.02);
  \fill[gray!4] (4.48,0.22) rectangle (9.52,4.00);
  \fill[blue!4] (4.90,0.56) rectangle (8.67,3.72);

  \begin{scope}
    \clip (4.48,0.22) rectangle (9.52,4.00);
    \begin{scope}[rotate around={-30:(7.15,2.13)}]
      \draw[AlbGray!30, dashed, line width=0.65pt] (5.30,0.55) rectangle (9.25,3.72);
      \draw[AlbGray!24, line width=0.45pt] (5.30,1.25) -- (9.25,1.25);
    \end{scope}
  \end{scope}

  \begin{scope}[rotate around={-30:(6.90,2.10)}]
    \path[fill=gray!55, draw=gray!70]
      (6.50,1.52)
      .. controls (6.20,1.82) and (6.24,2.38) .. (6.62,2.61)
      .. controls (6.86,2.75) and (7.12,2.69) .. (7.29,2.50)
      .. controls (7.66,2.23) and (7.61,1.72) .. (7.29,1.48)
      .. controls (7.05,1.30) and (6.72,1.32) .. cycle;
    \path[fill=AlbGreen, fill opacity=0.34, draw=AlbGreen, line width=0.75pt]
      (6.50,1.52)
      .. controls (6.20,1.82) and (6.24,2.38) .. (6.62,2.61)
      .. controls (6.86,2.75) and (7.12,2.69) .. (7.29,2.50)
      .. controls (7.66,2.23) and (7.61,1.72) .. (7.29,1.48)
      .. controls (7.05,1.30) and (6.72,1.32) .. cycle;
    \draw[AlbGray!80, line width=0.85pt] (6.92,2.60) -- (6.85,2.82);
  \end{scope}
  \draw[applebox] (6.15,1.28) rectangle (7.65,2.90);
  \node[applelabel] at (6.15,2.90) {apple};

  \begin{scope}[rotate around={-30:(9.18,2.07)}]
    \path[fill=gray!55, draw=gray!70, rounded corners=1.0pt]
      (8.92,1.62) rectangle (9.38,2.42);
    \draw[gray!70, line width=3.2pt]
      (9.36,2.24) arc[start angle=90,end angle=-90,x radius=0.23cm,y radius=0.27cm];
    \path[fill=AlbPurple, fill opacity=0.34, draw=AlbPurple, line width=0.7pt,
      rounded corners=1.0pt] (8.92,1.62) rectangle (9.38,2.42);
    \draw[AlbPurple, opacity=0.55, line width=3.2pt]
      (9.36,2.24) arc[start angle=90,end angle=-90,x radius=0.23cm,y radius=0.27cm];
  \end{scope}
  \draw[mugbox] (8.39,1.25) rectangle (9.43,2.78);
  \node[muglabel] at (8.39,2.78) {mug box};
  \fill[AlbPurple, fill opacity=0.18] (8.39,1.25) rectangle (8.67,2.78);

  \draw[outputframe] (4.90,0.56) rectangle (8.67,3.72);
  \node[fill=AlbBlue, text=white, font=\scriptsize\bfseries,
    inner xsep=3pt, inner ysep=1.4pt, anchor=north west]
    at (4.90,3.72) {output frame};
  \node[font=\scriptsize, text=AlbPurple, align=center]
    at (9.00,0.62) {box sliver\\inside};
  \draw[AlbPurple, line width=0.55pt] (8.88,0.88) -- (8.57,1.27);

  \draw[flow] (9.67,2.20) -- (10.40,2.20);
  \node[font=\scriptsize\bfseries, text=AlbBlue, align=center]
    at (10.03,2.58) {filter};

  \draw[panel] (10.52,0.55) rectangle (14.34,3.90);
  \fill[blue!5] (10.54,1.28) rectangle (14.32,3.88);
  \fill[brown!12] (10.54,0.57) rectangle (14.32,1.28);
  \draw[AlbGray!28, line width=0.45pt] (10.54,1.28) -- (14.32,1.28);

  \begin{scope}[rotate around={-30:(12.20,2.08)}]
    \path[fill=gray!55, draw=gray!70]
      (11.80,1.50)
      .. controls (11.50,1.80) and (11.54,2.36) .. (11.92,2.59)
      .. controls (12.16,2.73) and (12.42,2.67) .. (12.59,2.48)
      .. controls (12.96,2.21) and (12.91,1.70) .. (12.59,1.46)
      .. controls (12.35,1.28) and (12.02,1.30) .. cycle;
    \path[fill=AlbGreen, fill opacity=0.34, draw=AlbGreen, line width=0.75pt]
      (11.80,1.50)
      .. controls (11.50,1.80) and (11.54,2.36) .. (11.92,2.59)
      .. controls (12.16,2.73) and (12.42,2.67) .. (12.59,2.48)
      .. controls (12.96,2.21) and (12.91,1.70) .. (12.59,1.46)
      .. controls (12.35,1.28) and (12.02,1.30) .. cycle;
    \draw[AlbGray!80, line width=0.85pt] (12.22,2.58) -- (12.15,2.80);
  \end{scope}
  \draw[applebox] (11.45,1.26) rectangle (12.95,2.88);
  \node[applelabel] at (11.45,2.88) {apple};
\end{tikzpicture}%
}